# Fabricated Pictures Detection with Graph Matching

Binrui Shen
Laboratory for Intelligent Computing and Finance Technology
Xi'an Jiaotong-Liverpool University
Suzhou, Jiang Su, China
binrui.shen19@student.xjtlu.edu.cn

Qiang Niu
Laboratory for Intelligent Computing and Finance Technology
Xi'an Jiaotong-Liverpool University
Suzhou, Jiang Su, China
qiang.niu@xjtlu.edu.cn

Shengxin Zhu
Laboratory for Intelligent Computing and Finance Technology
Xi'an Jiaotong-Liverpool University
Suzhou, Jiang Su, China
shengxin.zhu@xjtlu.edu.cn

## ABSTRACT

Fabricating experimental pictures in research work is a serious academic misconduct, which should better be detected in the reviewing process. However, due to large number of submissions, the detection whether a picture is fabricated or reused is laborious for reviewers, and sometimes is irrecognizable with human eyes. A tool for detecting similarity between images may help to alleviate this problem. Some methods based on local feature points matching work for most of the time, while these methods may result in mess of matchings due to ignorance of global relationship between features. We present a framework to detect similar, or perhaps fabricated, pictures with the graph matching techniques. A new iterative method is proposed, and experiments show that such a graph matching technique is better than the methods based only on local features for some cases.

## 1. INTRODUCTION

One kind of academic misconduct is that some authors reuse their experimental pictures by fabricating original ones. Even publications in esteemed journals have been reported to reuse fabricated pictures. Although most duplicates were not fraudulent or malicious manipulations, they did misrepresent results [15]. In this paper, we will show that such kind of academic misconduct may be detected and avoided at the submission stage by using the state-of-art computational vision technology and artificial intelligence.

The main technology used here is the so-called graph matching which has a wide application in computational vision and pattern recognition. In recent years, it has been used in fingerprint recognition [13], face recognition [18] and face authentication [7]. The aim of graph matching is to find a correspondence between nodes of two graphs. To show similarity between two pictures, we first transform two pictures into graphs respectively, then two graphs are compared by making links between their nodes, such links can be used to illustrate similarity between the two graphs and their corresponding pictures.

Graph matching methods can be classified into two major categories: exact matching and inexact matching [4]. Exact matching requires edge-preserving, which makes constraints strict. Tree search [9] and VF2 [5] are efficient methods to match two graphs exactly. By comparison, inexact matching methods are more flexible for some real-world problems. Tree search can also be applied in inexact graph matching [1]. However, faster approaches are usually based on continuous relaxation, which can transform a discrete graph matching problem into a continuous one. Typical examples of this kind include graduated assignment [6] and projected fixed-point method [12]. Recent advancement in the fundamental connections between such discrete graph theory and continuous problems has been established in [3], that's why we focus on continuous relaxation method.

## 2. TRANSFORMATION OF IMAGE

When two pictures are transformed into graphs, the detection of similarity between two pictures can be considered as a graph-matching problem. In this section, we will first explain how to transform an image into a graph.

### 2.1 Image Features Extraction

Every image can be characterized by its features. Features here are referred to as some prominent characteristics which should have some local invariant properties such as translation invariance, scale invariance and rotation invariance. Such features can be extracted by the Scale-Invariant Feature Transform (SIFT) method [11], which is detailed in [10]. To reduce computational cost, this method uses a close approximation to the Laplacian of Gaussian which is also used to detect edges in pictures [2]. Figure 1 illustrates key points of a Liverpool bird which is a part of the logo of Xi'an Jiaotong-Liverpool University (XJTLU).

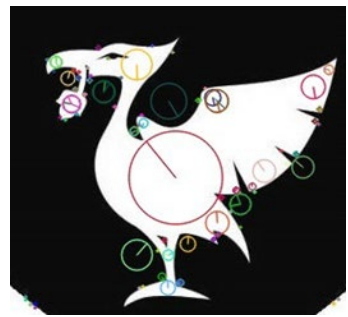

**Figure 1: Key points of a Liverpool bird**

### 2.2 Construction of Graph

After obtaining key points, one can set them as vertices of an undirected graph $G$. This set of vertices is denoted as $V$. Then we can link these nodes to get edges of $G$. The set of edges $E \subset V \times V$. At this stage, the image has been transformed into an unweighted complete graph without self-loop. The adjacency matrix of $G$ is denoted as $A$, and the size of $A$ is $n \times n$, wher $A_{ij}$ is the Euclidean distance between two nodes in this paper, in general it can be a general weight of the edge $E_{ij}$. What's more, all nodes' features form a feature matrix $F \in$

$R^{n \times p}$, where $F_i$, the $i_{th}$ row of $F$, is a p-dimensional vector which describes the feature of the $i_{th}$ node, by default, p = 128. Similarly, we can obtain a graph $\tilde{G}$ from another picture. The adjacency matrix and feature matrix of $\tilde{G}$ are denoted as $\tilde{A}$ and $\tilde{F}$ respectively.

## 3. GRAPH MATCHING

The matching relation can be represented as a permutation matrix $M$, where $M_{ij} = 1$ means the $i_{th}$ node of $G$ matches the $j_{th}$ node of $\tilde{G}$. For simplicity, we assume that the number of nodes of two graphs are equal to $n$, and the problem of graph matching with different size will be discussed later.

### 3.1 Objective function

To assess degree of matching of two graphs, a quantitative criterion is proposed in [19]:

$$\left\| A - M\tilde{A}M' \right\|_F^2 + \lambda \left\| F - M\tilde{F} \right\|_F^2, \tag{1}$$

where $\lambda$ is a regularization parameter, and $\|\cdot\|_F$ is the Frobenius matrix norm ($\|F\|_F^2 = \text{tr}(F^T F)$). The first term is the discrepancy between edges and the second term is the discrepancy between vertices' feature, then (1) is a total discrepancy. A small discrepancy implies a good matching. What's more, if two graphs are *isomorphic*, there exists a perfect matching which makes the discrepancy (1) zero. Therefore, the graph matching problem is transformed into the problem of minimizing the discrepancy (1)

$$\min_M \frac{1}{4} \left\| A - M\tilde{A}M' \right\|_F^2 + \lambda \left\| F - M\tilde{F} \right\|_F^2, \tag{2}$$

$$s.t. \quad M\mathbf{1} = \mathbf{1}, M'\mathbf{1} = \mathbf{1}, M \in \{0,1\}^{n \times n},$$

where the constraints enforce that $M$ is a permutation matrix. The coefficient 1/4 is chosen such that the formula can be transformed into a standard optimization problem [12]:

$$\max_M \frac{1}{2} tr(M'AM\tilde{A}) + \lambda tr(M'K), \tag{3}$$

$$s.t. \quad M\mathbf{1}=\mathbf{1}, M'\mathbf{1}=\mathbf{1}, M \in \{0,1\}^{n \times n},$$

where $K = F\tilde{F}'$.

### 3.2 Discrete to Continuous Optimization

Due to too many local minimums, the discrete optimization problem in (3) is usually NP-hard. To avoid getting trapped in a local minimum, the discrete problem (3) is transformed into a continuous one by relaxing the domain of the original problem onto the space of *doubly stochastic matrices*:

$$\max_M \quad \frac{1}{2} tr(M'AM\tilde{A}) + \lambda tr(M'K), \tag{4}$$

$$s.t. \quad M\mathbf{1} = \mathbf{1}, M'\mathbf{1} = \mathbf{1}, M \geq 0.$$

At this stage, it is a quadratic concave minimization problem on the space of doubly stochastic matrices. The $M$ in (4) can roughly be thought as a continuous matrix for a permutation matrix. After obtaining the final doubly stochastic matching matrix with Algorithm 1 in next subsection, the Hungary algorithm is used to transform the doubly stochastic matrix back to a permutation matrix [8].

### 3.3 Method for Continuous Optimization

The continuous optimization problem (4) can be solved by the projected fixed-point method proposed in [12]. Note that for

$$f(M) = \frac{1}{2} tr(M'AM\tilde{A}) + \lambda tr(M'K), \tag{5}$$

the gradient is:

$$\nabla f(M) = AM\tilde{A} + \lambda K.$$

Then the projected fixed-point method is defined as:

$$M^{(t+1)} = (1-\alpha)M^{(t)} + \alpha P(\nabla f(M^{(t)})), \tag{6}$$

where $\alpha$ is a controlling parameter for step size. $P(\cdot)$ is a projection function which aims to find a closest doubly stochastic matrix to a given matrix. Such a projection is the key to continuous relaxation method. A popular projection method is the so-called *softmax-Sinkhorn method* which will be detailed in next subsection. The projected fixed-point algorithm is shown in Algorithm 1.

---
**Algorithm 1** General Projected Fixed-Point

**Require:** $A_1, A_2, K, \lambda, \alpha$
**Ensure:** $M$
1: *Initial N*
2: **while** N not converges **or** *iterations* $< I_1$ **do**
3: $\quad N = (1-\alpha)N + \alpha P(A_1 N A_2 + \lambda K)$
4: **end while**
5: *Discretize N to obtain M*
6: **return** $M$

---

### 3.4 Softmax-Sinkhorn Method

The Softmax-Sinkhorn method consists of two parts: softmax method and Sinkhorn method [6]. Softmax is a technique to make all elements of matrix positive and increase discrepancy of elements:

$$N = \frac{\exp(\beta N)}{\sum_{ij} \exp(\beta N_{ij})},$$

where $\beta$ is a parameter to control the separate effect. Sinkhorn method is used to find the closest doubly stochastic matrix to a given matrix whose all elements are positive [16]. The doubly stochastic matrix can be obtained by an alternating iterative process of normalizing the rows and columns of the given matrix N [17]:

$$N_{ij} = \frac{N_{ij}}{\sum_i N_{ij}},$$

$$N_{ij} = \frac{N_{ij}}{\sum_j N_{ij}}.$$

The Softmax-Sinkhorn algorithm is described in Algorithm 2.

**Algorithm 2** Softmax-Sinkhorn

1: **function** $Sinkhorn(N, \beta)$
2:     $N = \frac{exp(\beta N)}{sum(exp(\beta N))}$
3:     **while** N not converges **do**
4:         $N_{ij} = \frac{N_{ij}}{\sum_j N_{ij}}$
5:         $N_{ij} = \frac{N_{ij}}{\sum_i N_{ij}}$
6:     **end while**
7:     **return** $N$
8: **end function**

## 3.5 Graduated Softmax-Sinkhorn Projected Fixed-point

For the graph matching problem with different sizes, one first should make these two graphs comparable [6][12] via a *slack matrix*: assume sizes of $G$ and $\tilde{G}$ are $n_1$ and $n_2$, where $n_1 > n_2$, we expand $M$ of size $n_1 \times n_2$ to a *slack matrix* of size $n_1 \times n_1$ in the projection process.

In experiments, we find that Algorithm 1 with Softmax-Sinkhorn projection is easy to converge to a local minimum, if we fix $\beta$ as a constant. To enhance the robustness of Algorithm 1, we use a graduated assignment method proposed by [6]: increase the value of $\beta$ after each run of the projected fixed-point algorithm gradually. In this graduated projected fixed-point method, three parameters must be determined to realize increasing $\beta$: initialization value $\beta_0$, maximum value $\beta_m$ and increasing rate $\beta_r$. The algorithm, Graduated Softmax-Sinkhorn Projected Fixed-Point (GSSPFP) algorithm, is shown in Algorithm 3.

**Algorithm 3** Graduated Softmax-Sinkhorn Projected Fixed-Point

**Require:** $A_1, A_2, K, \alpha, \lambda, \beta_0, \beta_r, \beta_m$
**Ensure:** $M$
1: $Initial\ N, N'$
2: $\beta = \beta_0$
3: **while** $\beta < \beta_m$ **do**
4:     **while** N not converges **or** $iterations < I_1$ **do**
5:         $N'(1:n_1, 1:n_2) = A_1 N A_2 + \lambda K$
6:         $N' = Sinkhorn(N', \beta)$
7:         $N = N'(1:n_1, 1:n_2)$
8:     **end while**
9:     $\beta = \beta \times \beta_r$
10: **end while**
11: $Discretize\ N\ to\ obtain\ M$
12: **return** $M$

## 4. Experiments

In this section, we compare the performance of graph matching based on GSSPFP and the performance of the feature matching method based on the Fast Library for Approximate Nearest Neighbors (FLANN). FLANN can directly give each feature point $i$ in $G$ the first $k$ similar candidate matching points in $G$ without consideration of edges ($k$=2 in this set of experiments). Such a point matching method is based solely on matching local feature points extracted by the SIFT.

    To be fair, we extract the same feature points from pictures for two matching methods. For the fixed-point iteration part, the convergence criteria is $max(|N^{t+1} - N^t|) < \epsilon_1$ or iteration number is more than $I_1$. The convergence criteria is $\|N^{t+1} - N^t\|_F < \epsilon_2$ in Sinkhorn iterative. As to choosing the regularization parameter, according to [12], the result is not sensitive to $\lambda$. For simplicity, we just choose $\lambda = 1$ in this paper. For experiments in section 4.1, the following values for the constants graduated increasing process are used: $\beta_0 = 10^{-6}$, $\beta_r = 1.2$ and $\beta_m = 5 \times 10^{-6}$. Due to the small size of picture in section 4.2, the values of $\beta_0$ and $\beta_m$ are $10^{-5}$ and $5 \times 10^{-5}$.

The relative position of a feature point $i$ among other points can be represented by $A_i$ which is the $i_{th}$ row of adjacency matrix $A$. Then the distance between the relative position of point $i$ and its matching point $\tilde{\iota}$ is $\|A_i - \tilde{A}_{\tilde{\iota}}\|_F$. The sum of all nodes' relative distance is $\|A - M\tilde{A}M'\|_F$, which can be considered as a positional discrepancy for a matching. Similarly, $\|F - M'\tilde{F}\|_F$ can measure the discrepancy between vertices' features. By combining the two terms, we get a matching error $\|A - M\tilde{A}M'\|_F + \|F - M'\tilde{F}\|_F$ for a matching $M$. This metric is used to compare two matching methods.

### 4.1 Designed Examples
#### 4.1.1 Wall Pictures
A set of 6 wall pictures from the public image dataset [20] are used as our benchmark problems, because they have many small similar features. Without global information, the results of the point matching method may be messy. We match the images in order, i.e. image 1 vs. image 2, image 3 vs. image 4, image 5 vs. image 6 and so on. Three matching results of them are shown in Figure 2, 3 and 4. The green (yellow) lines represent links between two points.

In this set of experiments, features of each image are very similar, which causes that the point matching method do not find the similarity between pictures clearly. When the information of edges is used by the graph matching method, the matchings results are clear. Table 1 also shows that the graph matching method achieve a lower matching. The matching error of five matchings are listed in Table 1. The error consists of the edge discrepancy and the node discrepancy.

A simple feature selection used in [12] is applied in this set of experiments due to the complex features in wall pictures. It is performed as follows: calculate the similarity (inner product) between a feature of one picture and all features in another picture, then choose the top $T$ features. In addition, the number of points of wall pictures is 500, and the number of points is 100 in later experiments.

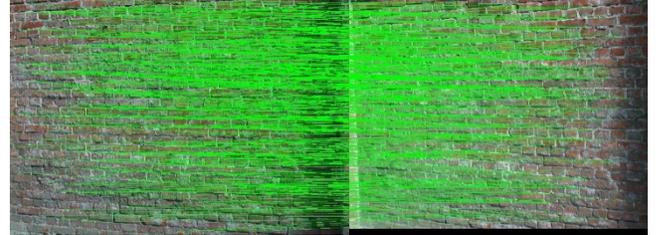
(a) Matching results by graph matching

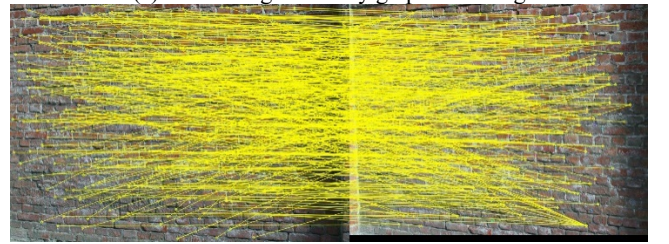

(b) Matching results by point matching

**Figure 2: Matching Wall Pictures Image 1 and Image 2**

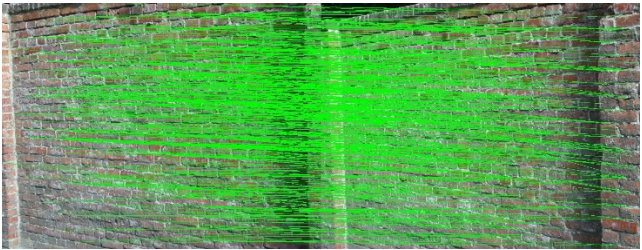

(a) Matching result by graph matching

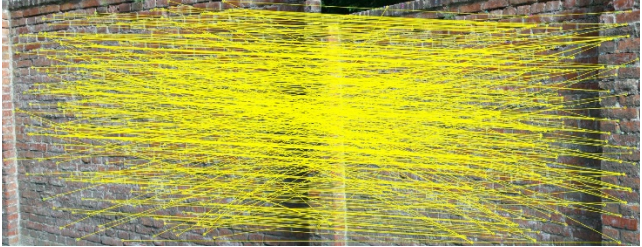

(b) Matching result by point matching

**Figure 3: Matching Wall Pictures Image 3 and Image 4**

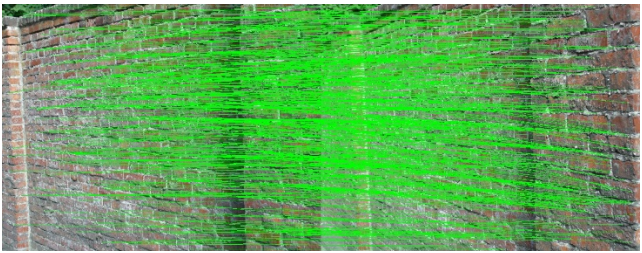

(a) Matching result by graph matching

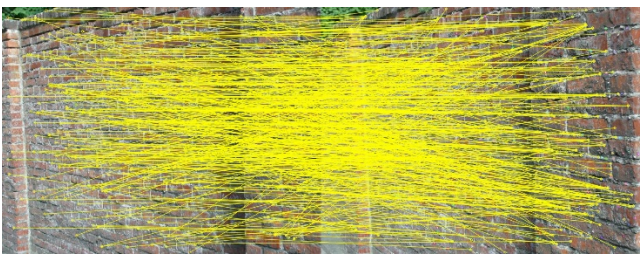

(b) Matching result by point matching

**Figure 4: Matching Wall Pictures Image 5 and Image 6**

**Table 1. Matching Error for Wall Pictures**

| Picture pair | FLANN | GSSPFP |
|---|---|---|
| Image 1 vs. Image 2 | 117750=111969 +5781 | 41997=33650 +8347 |
| Image 2 vs. Image 3 | 101200=96229 +4971 | 33775=25048 +8727 |
| Image 3 vs. Image 4 | 109766=104535 +5231 | 40194=31201 +8993 |
| Image 4 vs. Image 5 | 116485=111021 +5464 | 39986=31189 +8797 |
| Image 5 vs. Image 6 | 119924=113726 +6198 | 44999=35678 +9321 |

### 4.1.2 Photos of badge

In this set of experiments, we detect the similarity between real photos of XJTLU's badge. Firstly, a matching between a photo and its rotated form is shown in Figure 5. The results show that both methods are robust to rotational variations.

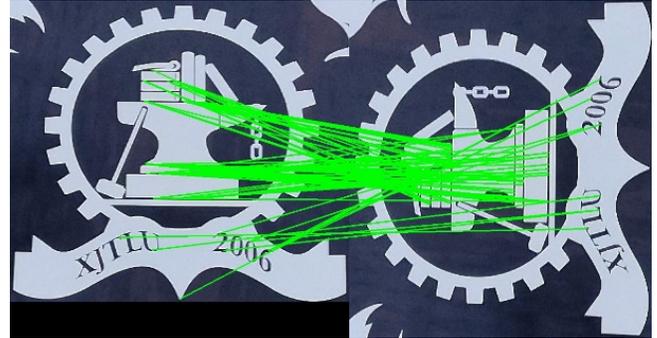

(a) Matching result by graph matching

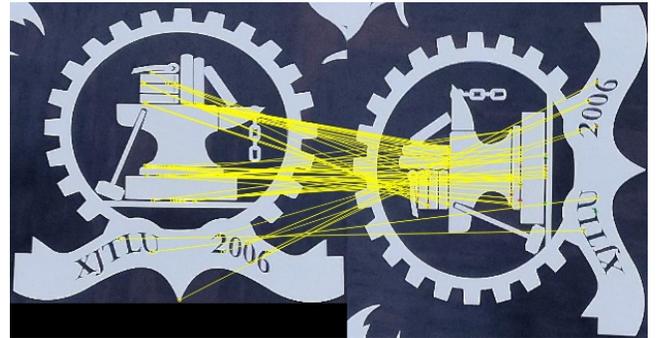

(b) Matching result by point matching

**Figure 5: Matching Rotated Badge Photos**

Figure 6 shows that when one photo is much blurred than another one, the point matching approach by FLANN does not work well while the graph matching approach still works.

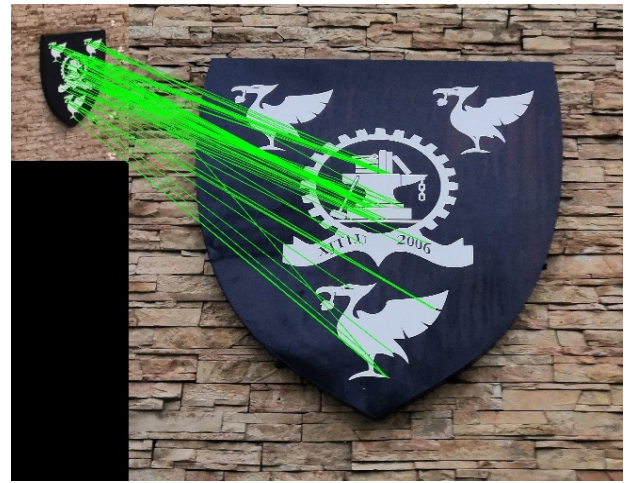

(a) Matching result by graph matching

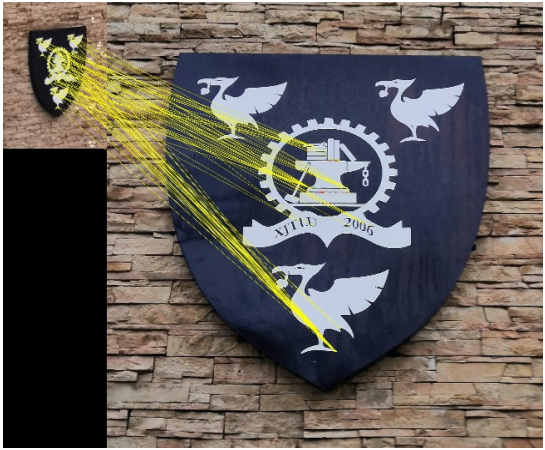

(b) Matching result by point matching

**Figure 6: Matching with a blurred badge photo**

Figure 7 illustrates that the graph matching still can detect the similarity of photos taken from different angles.

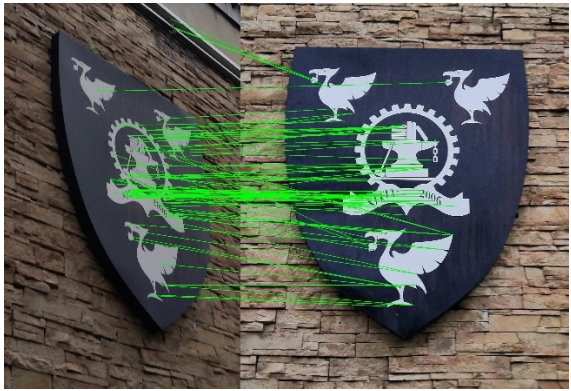

(a) Matching result by graph matching

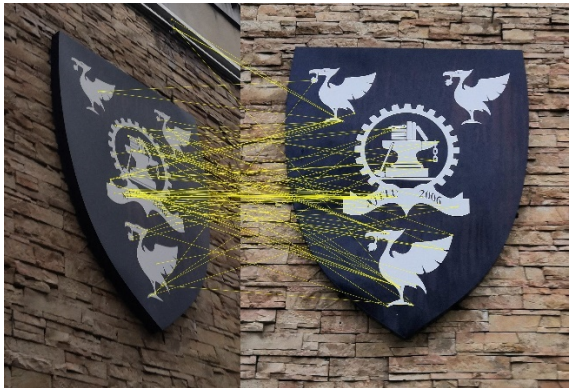

(b) Matching result by point matching

**Figure 7: Matching Photos from Different Perspective**

From the matching error in Table 2, we find that both methods can achieve a good performance when two photos are very similar. However, the graph matching approach works better if there exists difference between two photos, such as different degrees of clarity and photographing angles.

**Table 2. Matching error for Example 2**

| Examples | FLANN | GSSPFP |
|---|---|---|
| Matching with a rotated photo | 5215=4531+684 | 5478=4096+1382 |
| Matching with a blurred photo | 34781=31775+3006 | 22794=17887+4907 |
| Matching with a badge in profile | 50013=45472+4541 | 27596=21842+5754 |
| Example in 4.2 | 7698=5286+2412 | 4579=1473+3124 |

## 4.2 Fabricated Picture Detection Example

In this section, we give a real example of fabricated experimental picture detection. Figure 8 is an experimental image from a paper [[14], p.4 Figure A].

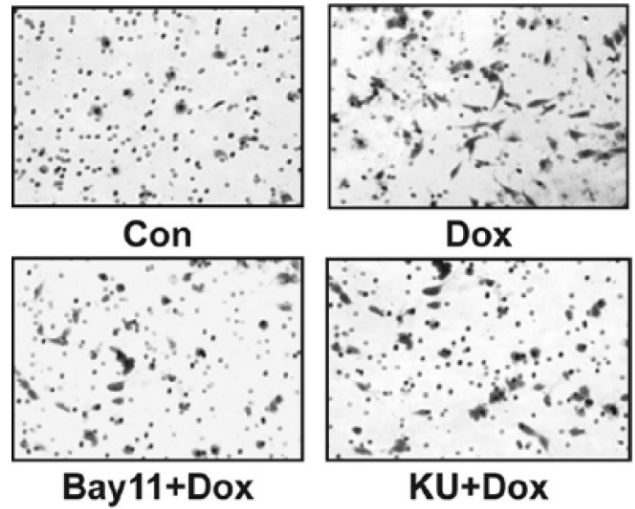

**Figure 8: Experimental picture from [13, p.4 Figure A]**

However, some scholars found that two of these sub-pictures were very similar ('Bay11 + DOX' and 'KU+Dox'). The bottom half of the third sub-picture is almost the same as the top half of the last sub-picture. To be clear, we use red windows to label the similar parts in Figure 9.

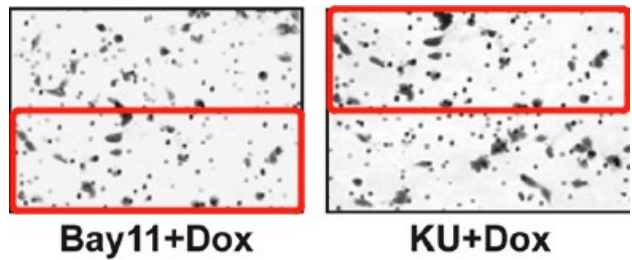

**Figure 9: Detection zones**

The relation of the two sub-pictures is shown in Figure 10. The matching error is listed in Table 2.

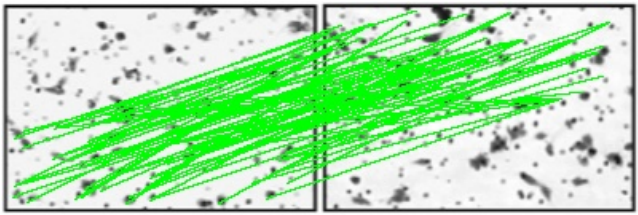
(a) Matching result by graph matching

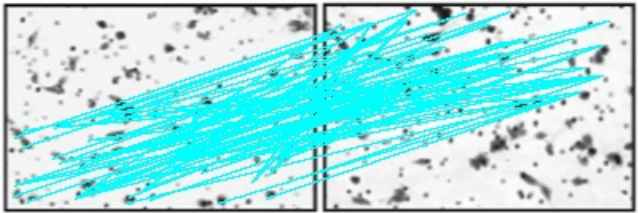
(b) Matching result by point matching

**Figure 10: Matching of sub-pictures**

## 5. DISCUSSION

This paper proposes a graph matching algorithm and applies it to detect fabricated pictures. The detective technique based on presented graph matching method is robust to variation including degrees of clarity and photographing angles. It can find the correspondence between an original image and its counterpart images, even though features of each image are extremely similar. Additionally, the correspondence can be shown clearly, which is helpful for users to judge whether a picture is fabricated from another one.

Although we propose a new frame of detective technique, some weakness remain in use. There are three parameters of $\beta$ need to be determined, which may cause a bad user experience. We will try to find a method to reduce the number of parameters by exploring the relation between $\beta$ and matching accuracy in the future. The second limit is that it cannot detect fabricated pictures automatically likes the Turnitin. It requires users to choose two pictures for detecting, which means it can only be an ancillary detective tool. In this direction, we need to make the program more intelligent.

## 6. ACKNOWLEDGMENTS

The research is supported by Laboratory of Computational Physics (6142A05180501), Jiangsu Science and Technology Basic Research Program (BK20171237), Key Program Special Fund in XJTLU (KSF-E-21, KSF-E-32, KSF-P-02), Research Development Fund of XJTLU (RDF-2017-02- 23), and partially supported by NSFC (No.11571002, 11571047, 11671049, 11671051, 61672003, 11871339).